\documentclass{article}
\usepackage[preprint]{tmlr}

\usepackage{graphicx}
\usepackage{booktabs}
\usepackage{amsmath,amssymb}
\usepackage{microtype}
\usepackage{url}

\newcommand{\bc}{\textsc{bc}}
\newcommand{\lo}{\textsc{local}}

\title{Human Interventions as Labels: Localizing, Steering, and (Sometimes) Editing the Computations Behind LLM Agent Failures}

\author{\name Cesare Zavattari \email cesare.zavattari@di.unipi.it \\
        \addr Dipartimento di Informatica, Universit\`a di Pisa, Pisa, Italy
        \AND
        \name Alessandro Tommasi \email alessandro.tommasi@di.unipi.it \\
        \addr Dipartimento di Informatica, Universit\`a di Pisa, Pisa, Italy
        \AND
        \name Giuseppe Prencipe \email giuseppe.prencipe@unipi.it \\
        \addr Dipartimento di Informatica, Universit\`a di Pisa, Pisa, Italy}

\def\month{07}
\def\year{2026}

\begin{document}

\maketitle

\begin{abstract}
When an LLM agent misbehaves, a human steps in --- and today that intervention is discarded or, at best, replayed as training data. We treat it instead as a \emph{measurement}: a label pointing at the internal computation that just failed. We study this idea in a deliberately small, fully controlled agent environment, on a stubborn failure --- asking a clarifying question when the request was already unambiguous --- using intervention labels issued by a rule-based overseer built to mimic a supervising human and validated against blind human judgment. All claims are restricted to \emph{eligible} models: those that both over-ask and sometimes act directly on matched unambiguous requests (a narrower window than one might expect; thirteen candidate families yielded five). Across those five families (Qwen2.5-3B, gemma-2-2b, OLMo-3-7B, Mistral-Nemo-12B, Phi-4) we report four findings. First, the ask-vs-act decision is \emph{causally localizable in every eligible family}: a direction learned from intervention labels flips 72--100\% of held-out over-asks at a definite residual-stream depth --- site and dose frozen on training episodes before one look at test --- with random directions under the identical procedure flipping at most one episode anywhere; the site's depth varies across families (55--81\% of the stack), its existence does not. Second, we observe \emph{two mechanistic signatures of over-asking}: in four families the model demonstrably knows the request is unambiguous --- the belief is decodable, causally inert, and orthogonal to the decision --- and asks anyway; in the fifth, the only well-calibrated over-asker, the rarer over-asks track genuinely different internal world-representations. The two signatures call for different fixes, and a probe readout separates them in our sample. Third, at runtime, steering \emph{gated on that readout} removes the majority of simulated oversight burden across seeds (pair-bootstrap 95\% intervals from $[0.59,0.94]$ to $[0.71,0.94]$, all excluding zero), while a prompt instruction makes the same burden worse. Fourth, consolidating the fix into the weights yields a map instead of a rule: on one model a LoRA confined to the causal band passes every criterion we fixed in advance, on three data-resample seeds --- and on two additional optimizer-seed retrains, one at the exact retention boundary --- while behavior cloning on identical data breaks retention on all three; on three other families the reverse holds, with the localized edit failing in three different diagnostic ways; on the fifth, the outcome crosses the pass/fail boundary we had fixed, across full-pipeline seeds. A geometric rule, written down in advance to predict which regime a model falls in, failed on two out-of-family tests; we report the failed rule and stop rather than fitting a replacement, and offer the five-family table as the constraint any future theory of localized editing must satisfy. Success throughout is scored by a budgeted overseer's burden; its intervention verdicts, and the headline intervention reductions, are validated blind by human annotators.
\end{abstract}

\section{Introduction}
\label{sec:intro}

Consider an LLM agent operating a small workspace. The user says: ``ship the draft addressed to Sam.'' There is exactly one draft addressed to Sam. The agent asks: ``Which draft do you mean?''

Nothing is broken, exactly. The final action, one turn later, is usually correct. But a human had to answer a question that the context already answered, and if the agent does this on most requests --- the models we study do it on 82--93\% of unambiguous ones --- the cost adds up. Someone supervising a fleet of such agents spends their attention on interventions that should not have been necessary. Oversight, not capability, becomes the bottleneck.

What do we do with those interventions today? Mostly nothing. Some pipelines replay them as fine-tuning data, treating the human correction as one more demonstration. We propose to treat it instead as an \emph{experimental result}. An intervention marks a precise moment where the model's computation went somewhere a supervisor had to undo. If we collect those moments and look inside the model at exactly those points, we should find the computation responsible --- and once found, we can try to change it, and \emph{only} it.

One honest scoping note before anything else, because the paper's title could be read as more than we did. The \emph{motivation} is real human oversight. The \emph{experimental labels} are issued by a versioned, rule-based overseer that watches episodes and intervenes the way a supervising human would --- a controlled, reproducible proxy for a human intervention log. The \emph{validity} of that proxy is checked separately: blind human annotation studies (\S\ref{sec:humans}) confirm that its verdicts, and the intervention reductions we report, agree with human judgment. We study human-intervention-\emph{shaped} oversight signal under laboratory control; deploying the same pipeline on naturally occurring human logs is the obvious next step, not something this paper claims.

The reframing suggests a different success criterion, too. Knowledge-editing papers score an edit by whether the model now answers a probe question correctly. An operations team would ask something else: \emph{did the overseer have to intervene less afterwards?} We adopt that standard and define a metric for it --- the Intervention Reduction Rate (IRR; not inter-rater reliability), the relative reduction of a budgeted overseer's burden, which combines watching costs and intervention costs (\S\ref{sec:setting}) --- and we anchor it in the quantities humans actually validated: full-attention failure counts accompany IRR where it is reported. This is the ``actionable'' standard recently articulated for interpretability evaluation \citep{orgad2026actionable}, applied literally.

We ran this program end to end on five eligible model families, with quantitative predictions written down before any edited model was evaluated. One headline hypothesis did not survive, and we consider its failure, under rules we fixed in advance, part of the contribution. The paper reports four findings:

\begin{enumerate}
\item \textbf{Intervention labels causally localize the ask-vs-act decision --- five for five among eligible families (\S\ref{sec:causal}).} Linear probes trained on intervention-labeled episodes read the upcoming decision essentially perfectly from mid-to-late layers; that alone proves little (under greedy decoding, a rich content baseline also predicts behavior), so the claim rests on the causal test: setting a \emph{single coordinate} --- the projection on the probe direction --- at the decision token, at the \emph{extrapolated} dose fixed in advance (one inter-class gap past the acting class), flips held-out over-asks into well-formed, correctly-targeted actions in every eligible family we instrumented: per-family point estimates 72--100\%, every family's Wilson 95\% lower bound above 61\% (frozen-selection protocol; the milder class-mean dose flips 12--20\% on four families and 75\% on Phi-4), while random directions under the identical procedure flip at most 1/74 anywhere. The site's \emph{location} varies widely (55--81\% of network depth); its existence and the sharpness of its onset do not.
\item \textbf{The decision separates from the model's ambiguity belief, revealing two signatures of over-asking (\S\ref{sec:signatures}).} In four families the belief representation is correct even in over-ask episodes, geometrically orthogonal to the decision axis, and causally inert in both directions: the model knows, and asks anyway (\textbf{H-caution}). The fifth family, Phi-4 --- the only well-calibrated over-asker in our sample --- shows the contrasting signature (\textbf{H-misrep}): its over-asks track genuinely different belief states. One family is not a taxonomy, but the contrast matters practically: ``improve the model's ambiguity detector'' is the natural fix, and for four of our five families it targets a component that is demonstrably not broken.
\item \textbf{Probe-gated steering reduces oversight burden where prompting increases it (\S\ref{sec:steering}).} A runtime policy that intervenes only when the two probes disagree --- world-says-clear $\wedge$ model-about-to-ask --- removes the majority of simulated oversight burden across three seeds (pair-bootstrap 95\% intervals $[0.59,0.94]$, $[0.68,0.94]$, $[0.71,0.94]$) with legitimate clarifying questions preserved, while a system-prompt instruction makes burden \emph{worse} on every seed. The cross-model twist (on gemma a ``question style'' direction outperforms the behavior direction, and the geometry says why) refines the claim: the specificity lives in the \emph{gate}; the best lever direction is model-dependent.
\item \textbf{Weight consolidation yields a five-family map --- one stable localized win, a strong imitation default, and two falsified predictors (\S\ref{sec:map}).} On Qwen, a LoRA confined to the causal band passes every criterion fixed in advance (\S\ref{sec:map}) on three seeds --- data/evaluation resamples sharing one training seed; two further retrains of the first dataset with fresh optimizer seeds also pass, one exactly at the retention boundary --- while behavior cloning on identical data (\bc, the interactive-imitation baseline, which we also call the imitation default) breaks retention on all three and introduces a failure mode the base model never had --- to our knowledge the first replicated case where mechanistic targeting beats the strong imitation default. It is also, in our sample, the only stable one: on three families the default wins and the localized edit fails in three different diagnostic ways; on Phi-4 the outcome crosses the criteria across full-pipeline seeds. We wrote down in advance a cheap geometric rule to predict the regime; it died on its first out-of-family test, and its revision, written down before the next models ran, died on the pre-declared discriminating model. Per the stopping rule we had set ourselves beforehand, we fit no third rule.
\end{enumerate}

A fifth result deserves mention outside the contribution list because it is a guardrail rather than a finding about internals: our \emph{first} candidate failure --- re-asking a question the user had already answered, at 45--67\% across three models --- turned out not to be in the weights at all. It vanished when the scaffold showed the agent its own previous question, and, closing the loop, the behavior-cloning edit later \emph{re-created} it through the weights (\S\ref{sec:notweights}). Surface behavior underdetermines the fix; cheap scaffold ablations should gate any mechanistic pipeline.

The decisive thresholds, predictions, and stopping rules were written down before the runs, with conventions declared before the models that made them matter were run (\S\ref{sec:prereg}); the harness is guarded by blocking invariants that caught five bugs before they could become conclusions (\S\ref{sec:collect}); and the judges were validated by humans before we trusted their numbers (\S\ref{sec:humans}).

\section{Related work}
\label{sec:related}

\paragraph{Activation steering and representation engineering.} Steering vectors \citep{turner2023actadd}, contrastive activation addition \citep{rimsky2024caa}, representation engineering \citep{zou2023repe} and conditional gating \citep{lee2025cast} established that single directions can control high-level behavior, and that a \emph{condition} on when to steer adds selectivity. We inherit the machinery and add two things: the \emph{label source} (oversight interventions on an acting agent, rather than curated contrastive prompts) and the \emph{metric} (reduction of future interventions rather than probe accuracy). Closest to our runtime component, recent work conditions inference-time steering on signals learned from automatic success/failure labels --- COAST in robotics \citep{miao2026coast}, conformal per-step probes for LLM agents \citep{padhi2026conformal}, probe-conditioned head interventions for calibration \citep{li2026pchi} --- and ClarifySAE \citep{petrova2026clarifysae} steers models \emph{toward} asking clarifying questions when instructions are genuinely ambiguous: the complementary half of our problem, which our retention criterion polices from the other side. None of these uses oversight signal as the label.

\paragraph{Knowledge editing and its ripple problem.} ROME \citep{meng2022rome} and MEMIT \citep{meng2023memit} edit facts --- a key--value association rewritten in one MLP --- which does not transfer to our target, a policy conditioned on the whole context; the field's persistent finding is that edits fail to \emph{propagate} \citep{cohen2024rippleedits,zhong2023mquake}, motivating circuit-aware editing \citep{yao2025cake,zhao2026mcircke}. Selective layer placement has also been studied for \emph{efficiency} \citep{xu2026layercards}. Our question is different: not whether a restricted edit is cheaper, but whether restriction to a \emph{causally implicated} band changes what the optimizer generalizes --- and our answer (\S\ref{sec:map}) is ``sometimes decisively for the better, more often for the worse, and no cheap measurement we tested predicts which.''

\paragraph{Learning from human interventions.} Interactive imitation learning \citep{ross2011dagger,kelly2019hgdagger} uses expert corrections --- up to full human takeovers --- as training data at the policy level; recent robotics work converts action-space corrective guidance into supervision for its noise-space policy, with impressive sample efficiency \citep{lu2026unisteer}. Nobody, as far as our verification found, has connected that signal to the model's \emph{internals}. We keep the strongest version of this line as our baseline: behavior cloning (\bc) on exactly the corrections our overseer produced, with exactly the fine-tuning recipe of our localized edit --- the comparison isolates the one variable we claim matters.

\paragraph{Agent failures and the knowing--doing gap.} The agent literature approaches our failure from both sides. \citet{min2025selfregulation} train agents to request interventions \emph{only when needed} under an explicit budget --- the mirror image of over-asking; BehaviorBench \citep{huang2025behaviorbench} frames agent behavior change as a model-editing task. Closest in spirit, the ``Lazy Agent'' failure motivating ASA \citep{wang2026asa} is a knowing--doing gap on tool use --- the model internally recognizes a tool is needed and fails to call it --- patched with training-free activation steering. Concurrently with our work, \citet{cheng2026knowingdoing} showed the correlational face of the same phenomenon inside the model: representations of tool \emph{necessity} and of the tool-\emph{call} action are both linearly decodable, yet their probe directions become nearly orthogonal exactly in the late-layer, last-token regime that drives the action. Our results converge with theirs from a different behavior and add the causal layer: the correct belief is not merely rotated away from the action --- intervening on it does nothing, in either direction --- and the dissociation is a \emph{variable}, not a constant (\S\ref{sec:signatures}). Independently, the WANDERING arc \citep{vicentino2026leverislate,vicentino2026circuitbreakers} localizes a different agentic decision (loop finalization) to a late causal band, with knowledge consolidating tens of layers before the action --- convergent evidence, from a different failure, that agentic decisions have late causal levers that the model's own knowledge does not automatically reach.

\paragraph{Actionability, auditability, and predictions made in advance.} A recent position paper argues interpretability should be evaluated by \emph{actionability} --- whether findings support validated decisions outside interpretability itself \citep{orgad2026actionable}. On methodology, the Predict--Intervene--Validate framework \citep{bugaud2026piv} formalizes what we practice --- falsifiable predictions frozen before test data, interventions with matched controls, out-of-distribution validation of the \emph{claim} --- and a pre-registered causal test of MoE modularity \citep{salomone2026moemodularity} recently demonstrated the same discipline, with most of its registered hypotheses dying in public; calls for auditable interpretability \citep{lan2026auditable} make the case general. Our contribution to this thread is an existence proof at LLM-agent scale on \emph{behavioral editing} (\S\ref{sec:prereg}, \S\ref{sec:map}).

\section{Setting: an agent, an overseer, and a failure worth editing}
\label{sec:setting}

\paragraph{The environment --- a causal testbed, not a simulation of deployment.} The measurements this paper runs on require three things real agent traces almost never give: ground truth about whether a request is ambiguous, matched contrasts that hold everything else fixed, and hundreds of balanced episodes. Our \texttt{clarify} environment buys those properties by being deliberately small. Each episode shows the agent a one-line task (``Ship the Sam draft now'') and a workspace of three items with attributes. The agent must reply in one line: \texttt{ACTION: <id>} or \texttt{ASK: <one question>}. Each scenario comes as a \textbf{twin pair}: the instruction is identical, word for word; only the workspace content differs, making the request either uniquely resolvable or genuinely ambiguous. Any other difference between twins is a confound, so the harness diffs the rendered prompts and refuses to run if anything but the item lines differs. The trade --- external validity for internal validity --- is made with open eyes: this environment is where causal claims are \emph{possible}; naturalistic replication is future work (\S\ref{sec:limitations}).

\paragraph{The failure, and who is eligible to have it.} In bf16 precision, the two models we studied first over-ask massively: at collection (300 pairs), 84\% and 90\% of unambiguous requests draw a clarifying question (Qwen2.5-3B-Instruct, gemma-2-2b-it). Of the three families added later, two over-ask at 88--90\% (OLMo-3-7B, Mistral-Nemo-12B) and one at 36\% (Phi-4; rates at screening and on evaluation base rows range 28--37\%). We call Phi-4 the one \emph{calibrated} over-asker in the sense that its over-ask rate sits far from saturation, leaving many direct actions to contrast against --- which \S\ref{sec:signatures} shows is not an incidental detail. This is $\tau_{\text{overask}}$, the primary failure type in the project's failure taxonomy. Crucially, a model is \textbf{eligible} for the study only if it produces both classes on unambiguous requests --- over-asks \emph{and} direct actions --- because every probe and every causal test needs a behavioral contrast. That window is narrow: of eleven candidate families screened, six sat at a degenerate corner (always-ask or never-ask), two could not hold the reply format, and three were eligible (Appendix~\ref{app:screening}). Every claim in this paper carries that scope.

\paragraph{The overseer, the metric, and what each is for.} Three roles must not be conflated. (i)~The \emph{labeling overseer} is a versioned, deterministic rule (rule-v2) that watches collection episodes and issues verdicts --- intervene or not, the failure type, and a canonical correction (\texttt{ACTION: b2}). The rule is the task's own definition applied to ground truth --- it knows which item the request denotes --- not a judge that guesses; it is the paper's stand-in for a supervising human because it is reproducible, auditable, and cheap enough to label thousands of episodes, and its verdicts are validated by humans (\S\ref{sec:humans}). (ii)~The \emph{evaluation overseer} is a separate budgeted process with attention limits and per-family trust dynamics --- it watches only a fraction of episodes, preferring families it trusts less, and its \textbf{burden} is
\[
  B = \frac{W + 2\,I}{N}, \qquad \mathrm{IRR} = \frac{B_{\text{base}} - B_{\text{edited}}}{B_{\text{base}}},
\]
where $N$ is the number of evaluation episodes, $W$ the number the overseer watched, and $I$ the interventions among the watched ones; failures on unwatched episodes cost nothing (parameters and trust dynamics in Appendix~\ref{app:overseer}). An edit's headline score is the IRR, the relative reduction of this burden. The two overseers apply the same deterministic rule to each episode --- there is no second judge --- so what the separation buys is not independence of judgment but a metric that is not the raw error rate the training labels came from: evaluation scenarios are fresh, only a watched fraction is scored, and watching and intervening are both costed. (iii)~\emph{Humans} validate both: blind annotation studies (\S\ref{sec:humans}) check the rule's verdicts ($\kappa = 0.90$--$0.97$ across annotators) and reproduce the intervention reductions; the budgeted-burden \emph{dynamics} themselves are an automated operational proxy, not directly human-validated, which is why full-attention failure counts accompany IRR where it is reported.

\section{Method: from interventions to a causal handle}
\label{sec:method}

The pipeline has four stages, and the ordering is the method's point: each stage's output is only trusted after the next stage's test. A fifth element --- predictions written down before the outcomes --- wraps the whole thing.

\subsection{Collect, with the prompt as part of the experiment}
\label{sec:collect}

We roll out the agent on fresh scenario pairs (300 pairs per seed, three seeds per model) and store \emph{everything}: the verbatim message list at every step, raw replies, parsed decisions, overseer verdicts. Three blocking checks run during collection: the replay invariant (generation stack and measurement stack must tokenize the re-rendered prompt identically, to the token; an external audit noted this guards tokenizer coherence rather than full rollout$\to$replay identity --- rollout and replay share the stored message lists and a pinned tokenizer, but the token ids themselves were not persisted; the harness now records them as a blocking check, applying to collections after this paper's), the pairwise prompt diff (\S\ref{sec:setting}), and a false-positive canary on the overseer (a guard for the LLM-judge overseer and for future policy versions; the deterministic rule cannot trip it). A run that fails any of them does not produce data.

These checks are not ceremony: together with the audits around them they caught five harness bugs across the project (Appendix~\ref{app:bugs}), two of which would have manufactured a conclusion. The one that matters most for \S\ref{sec:map}: on OLMo-3 the episode runner did not stop at the end-of-turn token, the next turn's role word got glued onto replies (\texttt{ACTION: d1} became \texttt{ACTION: d1user}), and the mis-parsed base model looked worse than it was --- flattering every edit compared against it. The \emph{base-row check} --- the evaluation harness must reproduce collection-time behavior on the unedited model before any edited number is read --- is an audit comparison rather than a blocking assert; it is what exposed the bug, the bug was fixed with a regression test, and the affected results below are from the repaired harness. After the repair, base-row over-ask rates agree with collection-time rates within 8 points on all 13 family$\times$seed evaluations. We disclose that we audited this bug knowing its effect had favored our method.

\subsection{Probe, then distrust the probe}
\label{sec:probes}

At the last prompt token --- the \emph{decision token}, where the model is about to commit to \texttt{ASK} or \texttt{ACTION}; it is the final formatting token of the chat template, and every intervention in this paper acts on that one position at that one step --- we train two logistic probes per layer: one for the upcoming \emph{behavior} (will it ask?), trained on over-asked vs.\ acted episodes among unambiguous requests, and one for the \emph{belief} (is the request ambiguous?), trained on the ground-truth ambiguity label. Splits are by pair, never by record, so twins cannot straddle train and test. Mandatory controls accompany every probe number: permuted-label canaries (must sit at chance, averaged over five permutations); for the belief probe, whose classes are the twins, a positional canary at the last \emph{instruction} token (twin activations are bit-identical there by construction, so any signal is a bug); and a rich content baseline --- bag-of-words plus target position, attribute values, and distractor counts --- that answers ``could a model predict this from the prompt text alone?''

The right reading of our probe results is deflationary: behavior probes reach AUC $\approx 1.0$ at mid-to-late layers on every eligible family, but under greedy decoding the enriched content baseline reaches 0.93--0.99 on Qwen too. Decodability is not the finding. The finding is what survives the causal test.

\subsection{One coordinate, both directions, three controls}
\label{sec:patching}

The behavior probe gives us a unit direction $d$ per layer --- its normalized weight vector, a weighted combination of all residual-stream coordinates. The causal test asks: if we set the activation's scalar projection on $d$ --- one coordinate in a basis that has $d$ as its first axis, ``single coordinate'' in that sense throughout --- to a value typical of the other class, touching nothing else in the thousands-dimensional stream, does the decision flip? Two dose choices fixed in advance, both computed on training episodes only: \textbf{class-mean} (move an over-asking episode's coordinate to the mean coordinate of episodes where the model acted) and \textbf{extrapolated} (overshoot past the acting class by one inter-class gap, testing dose-response).

We measure the decision as the logit difference between the \texttt{ASK} and \texttt{ACTION} opening tokens (one forward pass, no generation), then validate flips with full generation. The headline flip rates (Table~\ref{tab:causal}) are at the extrapolated dose; class-mean rates are reported alongside. Every patching experiment carries three controls with the \emph{identical} procedure, all run at the class-mean recipe. (1) \emph{Random directions}: along a random unit direction the two classes have nearly the same mean, so the recipe moves the activation little; this arm rules out the patching machinery itself, not a matched-norm shove, and displacement norms per arm were not recorded. (2) A \textbf{style direction}: the difference of mean activations under two system-prompt suffixes on the same unambiguous training scenarios --- one instructing the agent to ALWAYS ask, one to NEVER ask (Appendix~\ref{app:overseer}) --- a real axis with its own class gap that separates the instructed output mode without any ambiguity semantics (is this just ``make it terse'', or just ``perturb hard enough''?). At the dose-matched comparison, class-mean $d$ against style against random, Qwen L28 reads 15/74 vs 5/74 vs 0/74 and gemma L21 reads 9/76 vs 76/76 vs 0/76 --- the gemma number is the finding \S\ref{sec:steering} returns to. (3) Patching the \emph{belief} axis in both directions (the behavior axis is patched only from ask toward act, the direction the failure calls for); the double null on the belief axis is how \S\ref{sec:signatures} establishes causal inertness. The cosine between the behavior and style directions, used from \S\ref{sec:steering} on, is the dot product of the two unit vectors at the same site.

\subsection{Edit, always under the same evaluation harness}
\label{sec:edit}

The runtime intervention is deliberately minimal: a gate computed from the two probes (fire only if ``belief says unambiguous'' AND ``behavior says about to ask'') and a one-shot projection at the decision token; the continuation runs unpatched. The weight-level intervention trains a LoRA on the labeling overseer's \emph{corrections} plus retention examples (the model's own replies on clean episodes), in two variants differing in exactly one argument: whether trainable parameters are restricted to the causal band (\lo) or spread over all layers (\bc, the interactive-imitation baseline). All evaluations --- steering, edits, baselines --- go through one harness: fresh scenarios, the budgeted evaluation overseer of \S\ref{sec:setting} (distinct from the labeling overseer), a 50-item capability battery, and pass/fail thresholds fixed before the runs.

\subsection{Predictions before outcomes}
\label{sec:prereg}

Interpretability findings are notoriously vulnerable to post-hoc flexibility --- many sites, many thresholds, a narrative assembled after the numbers arrive \citep{bugaud2026piv,lan2026auditable}. Our protection was a discipline kept among ourselves and applied mechanically: \textbf{predictions before outcomes} (on every new family, the expected pass/fail of both arms per criterion was written down in the project notes, derived mechanically from the rule under test, before any edited model was evaluated --- on the first two families the notes followed the training launch by minutes, on the third they preceded it, so each prediction precedes its outcome, not every training start); \textbf{conventions declared early} (e.g., ``the peak is the first argmax'' was fixed before the model with a causal plateau, Phi-4, was run); \textbf{a stopping rule}, set together with the revised rule of \S\ref{sec:map} and reading, in translation: \emph{if the two rules diverge, the outcome decides which survives; if they agree and are both wrong, report and stop generalizing}; \textbf{symmetric reporting} (confirmations, falsifications, and the full screening all appear, with the prior predictions quoted next to outcomes in \S\ref{sec:map}). We do not call this pre-registration: nothing was filed with a registry or attested by a third party. These were commitments the authors made to themselves, in the project's own notes, and the reader has only our word that they came first. What the discipline buys is internal --- a prediction written down before the run cannot be quietly rewritten after it --- and \S\ref{sec:map} shows what that is worth: a falsification that could not be reinterpreted after the fact. Externally attested versions of the same practice have been argued for elsewhere \citep{bugaud2026piv,salomone2026moemodularity}.

\section{Results}
\label{sec:results}

\subsection{The signature exists and is causal --- in every eligible family}
\label{sec:causal}

\emph{(Per model: 300 pairs per seed --- three seeds on Qwen, OLMo-3, Nemo and Phi-4, one on gemma; direction patching on held-out over-ask episodes, replayed on prompts re-rendered from the stored verbatim message lists under the tokenization-identity check of \S\ref{sec:setting}. Flip rates below follow the frozen-selection protocol of Table~\ref{tab:causal} --- site and dose chosen on training over-asks, one evaluation on test, extrapolated dose.)}

The correlational picture is clean but, as argued, not probative: on every eligible family the behavior probe saturates (AUC 0.997--1.0) somewhere in the mid-to-late stack, and the belief probe shows a textbook computational curve --- chance at layer 0, a threshold in the middle layers, saturation before the behavior probe's peak.

\begin{figure}[t]
\centering
\includegraphics[width=\textwidth]{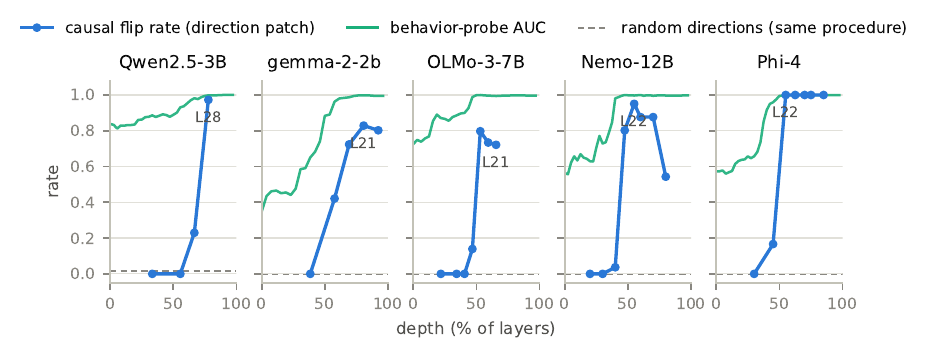}
\caption{Layer-wise causal profiles for the five eligible families: causal flip rate of the behavior-probe direction (extrapolated dose) against relative depth, with the behavior-probe AUC overlaid (it starts well above chance at layer 0, because prompt content is decodable from the start) and, dashed, the largest random-control flip over all sites at the class-mean recipe (one number, not a profile). The probe becomes readable well before the direction becomes causal; the onset of causal leverage is sharp in every family and at a different depth in each. Above the onset the effect is a plateau on gemma, OLMo-3 and Phi-4; on Qwen no site above L28 was tested.}
\label{fig:profiles}
\end{figure}

The causal picture is where the claim lives, and it repeats five times (Table~\ref{tab:causal}, Figure~\ref{fig:profiles}).

\begin{table}[t]
\centering
\caption{Causal localization of the ask-vs-act decision, per eligible family, under the frozen-selection protocol: site and dose are the arg-max, among the tested sites, of the same sweep run on \emph{training} over-asks only, then evaluated once on untouched test episodes ($n$ = test over-asks scored ASK-side by the first-token metric before the patch). Brackets: Wilson 95\% intervals on the extrapolated-dose flip rate; the class-mean column is the same site at the milder of the two doses fixed in advance. In the site column, the percentage is the site's depth in the stack and the note says where it sits on a causal plateau. Random directions: same site, same unit norm, same target-setting procedure, run at the class-mean recipe (there is no extrapolated random arm), so the dose-matched placebo comparison is the class-mean column against the random column; across \emph{all} sites and models the largest random-control flip is 1/74 (Qwen, at a causally null layer). On four families the train-side selection reproduces the site of the exploratory test-side sweep; on OLMo-3, whose causal region is a plateau (L17--L21, 72--80\%), it lands at L21.}
\label{tab:causal}
\footnotesize
\setlength{\tabcolsep}{4pt}
\begin{tabular}{lcccccc}
\toprule
model & layers & causal site (rel.\ depth) & flip (extrap.) & flip (class-mean) & $n$ & random \\
\midrule
Qwen2.5-3B-Instruct & 36 & L28 (78\%) & \textbf{97\%} [91--99] & 20\% & 74 & 0/74 \\
gemma-2-2b-it       & 26 & L21 (81\%) & \textbf{83\%} [73--90] & 12\% & 76 & 0/76 \\
OLMo-3-7B-Instruct  & 32 & L21 (66\%; plateau from L17) & \textbf{72\%} [61--81] & 20\% & 79 & 0/79 \\
Mistral-Nemo-12B    & 40 & L22 (55\%) & \textbf{95\%} [88--98] & 16\% & 81 & 0/81 \\
Phi-4 (14B)         & 40 & L22 (55\%; plateau to L34) & \textbf{100\%} [86--100] & 75\% & 24 & 0/24 \\
\bottomrule
\end{tabular}
\end{table}

The table makes two different points. Read the \emph{flip} column down: moving a single probe-derived coordinate at one site flips the large majority --- on Phi-4, all 24 --- of held-out over-asks into direct actions, while random directions under the same procedure at the same site flip zero episodes on any model (and at most one episode, 1/74 on Qwen, at any site at all). The profiles are sharp: on Qwen, layer 24 flips 23\% and layers 12 and 20 flip 0\%; on OLMo-3 the causal region is a short plateau (L17--L21, 72--80\%; the train-frozen selection lands at its late edge, which is why its table row reads 72\% where a test-side peak would have read 80\%); on Phi-4 the flip rate holds at 100\% along a plateau from layer 22 to 34 --- a structural difference \S\ref{sec:map} returns to. Full generation under the patch confirms the flips are not logit artifacts: on Qwen, 97\% flip, 99\% emit well-formed \texttt{ACTION:} lines, and 95\% act on the correct object; on gemma, all 65 episodes that flipped under generation acted on the correct object (65/65).

Now read the \emph{depth} column down: 78\%, 81\%, 66\%, 55\%, 55\%. The onset is always sharp and never at the same depth. Anyone hoping to reuse one model's coordinates on another --- the tempting shortcut interpretability keeps warning itself about --- would have failed three times out of five. Concurrent work reports the same double message for a different agentic decision (loop finalization consolidating in a late causal band; \citealp{vicentino2026leverislate,vicentino2026circuitbreakers}): the \emph{phenomenon} of a late causal lever travels; the \emph{address} does not.

\subsection{Two signatures of over-asking: knowing-without-doing, and genuine misreading}
\label{sec:signatures}

Why does a model that acts correctly 95\% of the time when patched ask the question in the first place? The natural suspect is misperception: it over-asks because it represents the request as ambiguous. In four families out of five, three measurements kill this hypothesis.

\begin{figure}[t]
\centering
\includegraphics[width=0.75\textwidth]{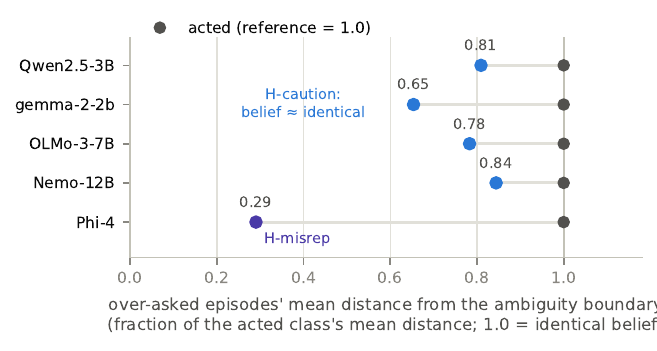}
\caption{The probe readout that separates the two signatures in our sample. For each family, the mean U-score of over-asked episodes as a fraction of the acted class's mean distance from the ambiguity boundary. In the four H-caution families, over-asked episodes carry essentially the same ``this is unambiguous'' belief as acted ones (0.65--0.84); on Phi-4 they sit far closer to the boundary (0.29): its over-asks track genuine internal uncertainty.}
\label{fig:useparation}
\end{figure}

Some notation first: call the belief probe's score of an episode its \textbf{U-score} --- the signed distance from the probe's decision boundary in logit units, positive side meaning ``ambiguous''. First, the readout: over-asked episodes sit firmly on the ``unambiguous'' side, nearly identical to acted episodes (Qwen, three seeds: mean U-score $-2.22$ vs $-2.65$; the same pattern on gemma, OLMo-3, and Nemo; Figure~\ref{fig:useparation}). Second, causality: on the two families where both faces were run, setting the belief coordinate to strongly-unambiguous removes almost no questions (at most 1/74 on Qwen, 0/76 on gemma), and pushing it to strongly-ambiguous on acted episodes induces almost none (0/13 on Qwen, 1/12 on gemma at the strongest dose --- small samples, with Wilson 95\% upper bounds of 23\% and 35\% respectively) --- the belief axis is causally disconnected in \emph{both} directions. Third, geometry: the belief and decision directions are near-orthogonal ($\cos \approx 0$) on every family tested. The model computes the fact correctly, stores it legibly, and the decision does not consult it. We call this signature \textbf{H-caution}: the failure lives in the coupling between knowing and doing. This is the causal-mechanistic face of the knowing--doing gap the agent literature documents behaviorally \citep{wang2026asa,min2025selfregulation}, and that concurrent work shows correlationally for tool use \citep{cheng2026knowingdoing}; our both-faces causal test adds the missing step --- the knowledge is not merely rotated away from the action; intervening on it does nothing.

Phi-4 breaks the pattern, and the break is what makes H-caution a finding rather than a truism. Phi-4 is the only calibrated over-asker in the sample (36\% over-ask at collection against 84--90\% elsewhere), and the only family where the belief readout genuinely differs between over-asked and acted episodes (mean U-score $-12$ vs $-41$: its over-asks sit far closer to the ambiguity boundary). Its over-asking tracks real internal uncertainty --- the signature we call \textbf{H-misrep}. With one family on that side of the contrast, we claim an observation, not a taxonomy: our five-family sample separates into four H-caution models and one H-misrep model, the probe readout separates the two signatures in this sample, and they call for different fixes --- H-misrep says ``improve the perception''; H-caution says the perception is fine and the decision coupling is the target. For most models we opened, the natural fix aims at the one component that is demonstrably not broken.

\subsection{Runtime steering: the gate carries the specificity}
\label{sec:steering}

\begin{figure}[t]
\centering
\includegraphics[width=0.7\textwidth]{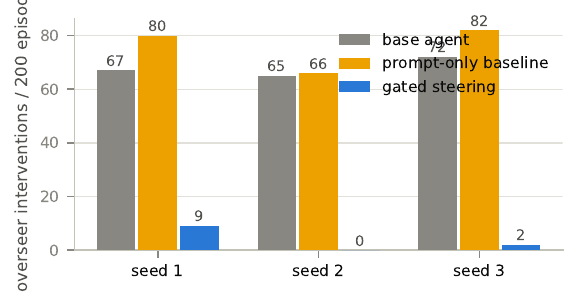}
\caption{Overseer interventions per 200 fresh episodes, per seed (Qwen). Gated steering removes most interventions; the prompt-only instruction \emph{increases} them on every seed.}
\label{fig:steering}
\end{figure}

The gated steering evaluation ran under criteria fixed in advance, on 200 fresh scenarios per seed, on Qwen, with three independently refit seeds. The pattern is identical across seeds (Figure~\ref{fig:steering}): overseer interventions drop from 67/65/72 to 9/0/2 --- these are interventions on the episodes the budgeted overseer chose to watch; counting also the failures it did not watch, total failures fall from 91/86/95 to 24/17/16 ($-74/-80/-83\%$), and it is full-attention counts of this kind, not the budgeted ones, that the blind human study of \S\ref{sec:humans} reproduces; simulated oversight burden falls by \textbf{73\%\,/\,91\%\,/\,87\%} at the default overseer seed --- point estimates that an external audit showed to be sensitive to the simulated overseer's own random seed (0.45--0.95 on seed~1 across 1{,}000 overseer seeds), so we report pair-cluster bootstrap 95\% intervals, which rerun the full dynamic-trust simulation per replicate: $[0.59,0.94]$, $[0.68,0.94]$, $[0.71,0.94]$, all excluding zero (the criterion as we had written it named a Wilson interval, which is undefined for a burden ratio; the bootstrap is the applicable analogue, and the deviation is declared here); legitimate clarifying questions barely move ($-2$ to $-4$ out of $\sim$93 --- and those few become proceed-when-ambiguous, 8$\to$11, 6$\to$8, 5$\to$9 of $\sim$100, an introduced error that the burden weights like an over-ask; \S\ref{sec:limitations}); the capability battery is unchanged (12 items at the time of these runs; the 50-item battery of \S\ref{sec:map} was introduced afterwards) and the gate essentially never fires off-target (two battery firings in 36 items across seeds). The two controls, run under the \emph{same gate}, are inert on Qwen. And the prompt-only baseline --- ``do not ask unnecessary questions'' in the system prompt --- makes things \emph{worse} on every seed: interventions 67$\to$80, 65$\to$66, 72$\to$82 (IRR $-20\%$, $-1\%$, $-11\%$). That contrast is the cleanest argument we have that internal intervention is not just a complicated way of writing an instruction.

The gemma replication forced a refinement. There, the style-direction control is \emph{not} inert: it flips 100\% at the causal layer and, in gated steering, outperforms the behavior direction (IRR 89\% vs 63\%). The geometry explains it: on gemma the style and behavior directions share half their alignment ($\cos$ 0.48--0.53 at causal layers) while on Qwen they are nearly orthogonal (0.05--0.12). So the claim that survives cross-model is not ``the behavior-probe direction is special''; it is that \textbf{selectivity lives in the gate} --- which fires identically for every direction and never off-target --- while the best lever direction is model-dependent. This changes how one should build such systems: invest in the gate; sweep the lever.

A natural rejoinder is that the entanglement might be an artifact of our dense probe basis --- perhaps a sparse dictionary would separate what the probes cannot. Gemma conveniently has one: we repeated the analysis with a Gemma Scope JumpReLU sparse autoencoder \citep{lieberum2024gemmascope} (16k features, residual stream, layer 21 --- the causal peak), under gates fixed in advance (transfer checks and one declared amendment; Appendix~\ref{app:sae}). The dictionary sharpens the picture rather than escaping it. \emph{Reading} is impressively sparse: ten features match the dense probe's near-perfect AUC on the behavioral label. But the entanglement reappears at feature resolution. The top-50 feature supports of the behavior and style axes share 19 features (0.15 expected for two unrelated sets) --- a correspondence one expects given the dense cosine of 0.56, so we read it as the dictionary failing to escape the overlap, not as independent evidence. The causal content is in the clamps, on 78 held-out over-asks at the class-mean dose: the style-orthogonal behavior-specific features flip nothing (0/78); the style-specific features flip nothing either (0/78) while moving the style projection by $-2.1$; the shared features flip 4/78 (Wilson 95\% [2, 12]\%, not separable from 0/78 at this sample size, Fisher $p = 0.12$) while moving style by $-6.5$; and the ten AUC-matching readers themselves --- clamped directly, in an arm added after an external audit found the original design never tested them --- flip 7/78 ([4, 17]\%, $p = 0.014$ against 0/78) at $-12.9$, twice the shared arm's style displacement (four of the ten sit inside the shared support). The random-feature arm moved nothing at all: its features are inactive on these episodes, so it is a null control of the clamping machinery, not a matched-magnitude placebo. What the arms support is this: on this model, at this site, the only sparse set that separates from zero is the readers, and it moves style the most; nothing that leaves style alone moves the decision --- consistent with the readers acting through their style-shared components, though an arm clamping the six non-shared readers alone would be needed to show it. This is a sharper form of the readable-versus-causal dissociation that concurrent work reports for an agentic stop-feature \citep{vicentino2026verdict}, here in dictionary space with the entanglement measured. Auto-interp labels put a memorable face on the numbers (the shared features that flip literally encode the two reply openers; the inert behavior-specific ones are textbook \emph{detectors} of choice and unclarity; Appendix~\ref{app:sae}), and we take the analysis as the geometric account restated at feature resolution, with the clamps as its causal content --- and with its caveats stated: class-mean dose only, one layer, one model, and per-arm displacement norms not recorded.

\subsection{Into the weights: the five-family map}
\label{sec:map}

The steering result leaves a practical objection: it needs probes and a hook at inference time. Can the correction be consolidated into the weights --- and does mechanistic localization earn anything over just fine-tuning on the corrections? This section's answer is a map, not a rule, and we give the map first.

\paragraph{The setup, everywhere the same.} Two LoRAs per model, trained on identical data (the labeling overseer's canonical corrections for intervened episodes; the model's own replies for clean ones --- retention) with identical hyperparameters. The \lo{} variant restricts trainable parameters to the causal band identified in \S\ref{sec:causal} (on Qwen: layers 24--31, 3.3M parameters). The \bc{} variant trains all layers (on Qwen: 15M parameters). Six criteria, fixed before any run. Four are pass/fail per arm: (a)~interventions at least halved; (b)~IRR $\geq 50\%$; (c)~legitimate clarifying questions drop by at most 10 of $\sim$100 (\emph{retention}); (d)~capability battery drops at most 5 points. Two compare the arms: (e)~localization must pay --- \lo{} passes (a)--(d) with at most a third of \bc's trainable parameters and either no worse a burden or smaller regressions on (c) and (d); (f)~the edited burden is reported next to gated steering's on the same scenarios, not as a pass/fail. On Qwen, (e) is met (3.3M against 15M parameters; burden 0.225 against \bc's 0.205, with \bc's retention loss the larger regression) and (f) reads 0.225 against gated steering's 0.385. One outcome term used below: \emph{proceed-when-ambiguous} --- acting without asking on a genuinely ambiguous request, the failure retention is designed to catch in its extreme form. Every evaluation's base row is checked against the model's collection-time behavior before any edited number is read (\S\ref{sec:collect}; agreement within 8 points on every family and seed).

\begin{figure}[t]
\centering
\includegraphics[width=\textwidth]{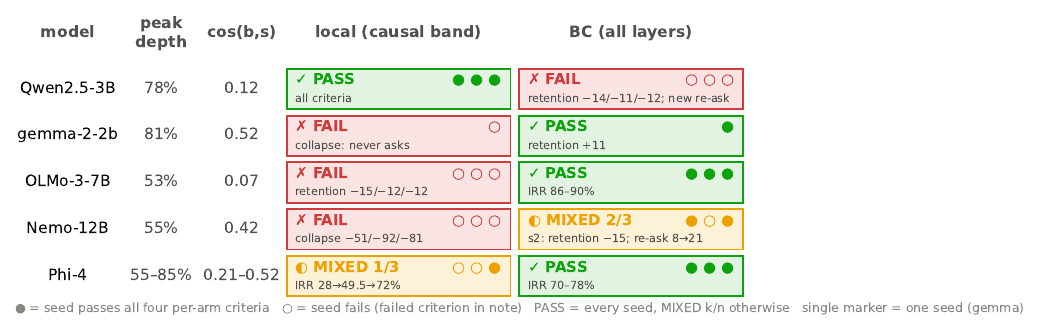}
\caption{The five-family editing map, every row at three full-pipeline seeds (fresh collection $\to$ both LoRAs $\to$ both evaluations per seed; gemma at one seed). Each cell: verdict against the four per-arm criteria (a seed passes only if it meets all four; a row is \textsc{pass} only when every seed does, \textsc{mixed} otherwise, with the count), per-seed outcomes, and the failed criterion. The depth column is the test-side peak of the exploratory sweep, the value the two rules were scored on (OLMo-3 53\%, Phi-4 55--85\% along its plateau); Table~\ref{tab:causal} gives the frozen-protocol sites. No cheap measurement we tested (cosine column, peak-depth column) predicts the \lo{} column --- both rules, written down in advance, were falsified out-of-family.}
\label{fig:map}
\end{figure}

\paragraph{The map.} Figure~\ref{fig:map} is the deliverable, and three durable statements summarize it. \textbf{(i) \bc{} from overseer corrections is the default to beat, and it usually isn't beaten} --- 9 of its 10 seed-arms outside Qwen pass everything --- but it is not universally safe: it fails retention on all three Qwen seeds and on one Nemo seed, and in those failures it also disturbs an unrelated behavior (re-asking an already-answered question) that the base model showed rarely or never. When \bc{} breaks something, it breaks it silently; check retention. \textbf{(ii) Localized editing is neither miracle nor mirage.} There is one replicated case --- Qwen, 3/3 seeds, all criteria --- where the aimed edit beats the default outright, with burden reductions of 84/85/86\% and the capability battery unchanged (0.84 base; 0.84/0.86/0.84 edited, on 50 items); and there are three families where it fails in three \emph{different, diagnostic} ways: unconditional collapse (gemma, Nemo), retention erosion (OLMo-3), and criterion-crossing instability (Phi-4, next paragraph). \textbf{(iii) No cheap measurement we tested predicts the regime.} Neither the decision axis's entanglement with style nor the causal peak's depth survived out-of-family prediction tests --- both were written down before the tests, both were falsified, and per the stopping rule we had set ourselves no third rule was fit. The map is the deliverable: any future theory of when interpretability-guided editing wins has ten arm-outcomes, each with its prior prediction, to be consistent with, and two dead rules it must not re-derive.

\begin{figure}[t]
\centering
\includegraphics[width=0.72\textwidth]{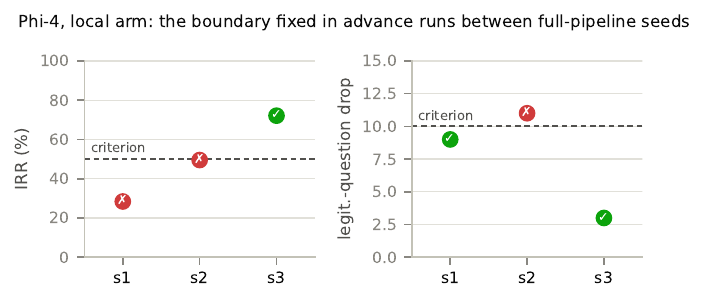}
\caption{Phi-4, \lo{} arm: the two criteria, fixed in advance, that it crosses, per seed. Green/check = criterion met, red/cross = missed.}
\label{fig:phi4}
\end{figure}

\paragraph{The Phi-4 row, unpacked.} Because a ``borderline'' verdict invites suspicion in both directions, Table~\ref{tab:phi4} shows the entire \lo{} arm, per seed (200 evaluation episodes per seed), and Figure~\ref{fig:phi4} plots the two criteria it crosses.

\begin{table}[t]
\centering
\caption{Phi-4, \lo{} arm, per full-pipeline seed. The boundary fixed in advance (IRR $\geq 50\%$, retention drop $\leq 10$) runs between the seeds. $B$ = burden, $I$ = interventions on watched episodes; failures = $I$ plus failures on unwatched episodes; clarify = legitimate clarifying questions, base$\to$edited.}
\label{tab:phi4}
\footnotesize
\setlength{\tabcolsep}{4pt}
\begin{tabular}{ccccccccc}
\toprule
seed & base $B$ & base $I$ & edited $B$ & edited $I$ & IRR & failures & clarify & verdict \\
\midrule
1 & 0.405 & 10 & 0.290 & 5 & 28\%   & 28$\to$10 & 100$\to$91 ($-9$)  & fail (b) \\
2 & 0.535 & 15 & 0.270 & 5 & 49.5\% & 38$\to$15 & 97$\to$86 ($-11$)  & fail (b, c) \\
3 & 0.430 & 10 & 0.120 & 0 & 72\%   & 33$\to$3 & 100$\to$97 ($-3$)  & \textbf{pass a--d} \\
\bottomrule
\end{tabular}
\end{table}

Note the base burdens, which are the denominators of IRR: Phi-4 is the only calibrated over-asker, so its baseline burden is already 0.40--0.54, against 1.40--1.54 on Qwen, 1.80 on gemma, 1.68--1.73 on OLMo-3 and 1.58--1.76 on Nemo; and about three quarters of Phi-4's burden is \emph{watching} (61 watched episodes against 10 interventions on seed 1), which an edit that removes failures reduces only slowly, because trust recovers gradually. A relative reduction on such a denominator swings widely with a handful of episodes --- while on total failures the \lo{} arm removes 64\%, 61\% and 91\% across the three seeds, above the bar on every seed. We write ``crosses the boundary across full-pipeline seeds'' rather than ``unstable model'' deliberately: each seed re-runs collection, training, and evaluation, so the variance includes all three. What the row establishes either way: the regime boundary this section maps does not just run between models; on Phi-4 it runs through the seeds of a single model --- one more reason a deterministic which-strategy-wins rule was never a live option.

\paragraph{How the map was obtained: two rules written down in advance, two falsifications.} The rest of this section is the audit trail --- how each row landed, in the order it happened, with the prediction we had written down next to each outcome.

\emph{Qwen, then gemma: the double dissociation.} On Qwen the \lo{} edit passes every criterion on all three seeds while \bc{} --- same data, same recipe, 4.5$\times$ the parameters --- reaches comparable headline burden (86/78/82\%) and fails the retention criterion every time ($-14/-11/-12$), while introducing the de-novo re-ask failure (15/9/11 episodes of 100). Spreading the parameters let the optimizer generalize ``ask less'' indiscriminately; confining them to the causal band did not. Then gemma inverted it: the \emph{restricted} edit over-generalizes catastrophically (never asks: 100/100 unambiguous episodes acted, 100/100 ambiguous too --- both ``perfect'' scores audited per our impossible-numbers protocol), while all-layers \bc{} is clean and even improves legitimate clarifying ($+11$). A perfect anti-diagonal --- a \textbf{double dissociation}, which rules out both trivial readings (localization generically helps; localization generically hurts).

\begin{figure}[t]
\centering
\includegraphics[width=0.72\textwidth]{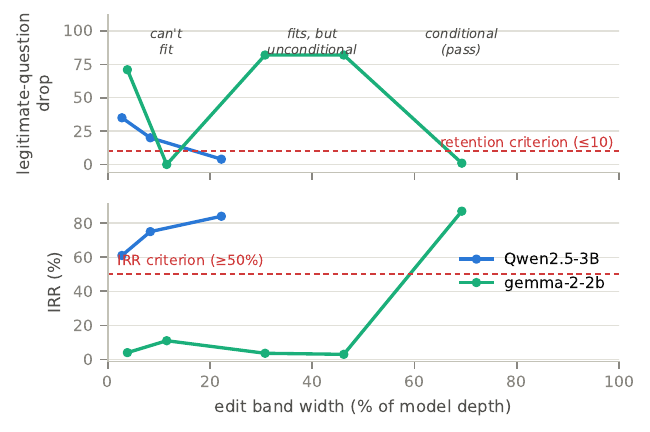}
\caption{Band-width sweep: retention (top) and IRR (bottom) against edit-band width as a fraction of model depth. Gemma shows the two ordered thresholds --- too narrow to fit, wide enough to fit but only unconditionally, wide enough for the conditional policy; Qwen passes at 22\% of depth.}
\label{fig:sweep}
\end{figure}

\emph{The rule.} The two models differ on a measurement we already had: $\cos(\text{behavior},\text{style}) \approx 0.12$ on Qwen vs $\approx 0.52$ on gemma. Hypothesis: entangled geometry needs distributed capacity; orthogonal geometry lets a narrow edit cut cleanly. A band-width sweep, with its predictions written down beforehand, supported the story within-model (Figure~\ref{fig:sweep}) and produced the section's sharpest concept: widening the band on gemma reveals \emph{two ordered thresholds} --- below the first (3 layers) the adapter cannot even fit the corrections; between them (8--12 layers) it fits them but generalizes as the crude unconditional rule (``never ask''); only above the second (18 of 26 layers, 69\% of depth) can it afford the \emph{conditional} policy (``act when unique, keep asking when not''). Qwen passes at 8 of 36 layers (22\%). Capacity for a correction and capacity for its \emph{conditionality} are different, ordered thresholds; the gap between them is the collapse zone. (One sweep sub-prediction died here and is reported: three layers do not suffice on Qwen, $-20$.) We wrote the cosine rule down as a mechanical prediction procedure for any new model.

\emph{Round one out-of-family: the rule dies.} OLMo-3 measured $\cos = 0.07$ --- Qwen-like, the cleanest disentanglement in the sample. Prediction written down beforehand: \lo{} passes, \bc{} fails retention. Outcome, on the repaired harness (\S\ref{sec:collect}): the exact inverse. \lo{} eroded legitimate clarifying (94$\to$79; everything else passing); \bc{} was spotless (94$\to$94, IRR 90\%). A model with Qwen's geometry behaved like gemma. What died is the \emph{prediction rule}; the Qwen/gemma dissociation is a pair of measurements and stood. One observation emerged from the wreckage --- OLMo-3's causal peak sits at 53\% of depth (L17, the test-side peak of the exploratory sweep, on which the \lo{} band L13--L20 was centred and on which the depth rule was scored; the frozen protocol adopted later lands at L21, 66\%, on the same plateau, Table~\ref{tab:causal}, a depth at which the revised rule would have made no prediction) versus 78--81\% for Qwen/gemma; perhaps a mid-stack band leaves too much downstream computation free to re-absorb the edit --- and per \S\ref{sec:prereg} it was written down \emph{before} the next two models ran, labeled post-hoc, side by side with the original rule and with the stopping rule.

\emph{Round two: the discriminating model.} Nemo-12B measured entangled \emph{and} mid-stack ($\cos$ 0.42, peak 55\%): both rules agreed \lo{} would fail, and it did, spectacularly (clarify $-51$; confirmation without discrimination). Phi-4 was the prize: a causal \emph{plateau} from 55\% of depth with the cosine rising along it (0.21 $\to$ 0.52) --- under the declared first-argmax convention, the original rule said \emph{\lo{} passes}, the revised rule said \emph{\lo{} fails retention}. Opposite predictions, one outcome. The outcome: \textbf{both wrong.} Retention held ($-9$, within criterion), killing the revised rule; burden reduction missed the bar (IRR 28\%), killing the original's ``passes everything''. \bc{} passed everything. The stopping rule we had set ourselves named two cases --- diverging rules, where the outcome picks a survivor, and concordant wrong rules, where one stops; Phi-4 produced the case it did not name, diverging and both wrong, and we applied its evident intent: no third rule; what would predict the regime is a question for theory, not for another variable adjusted on the same five points.

\emph{Replication.} The three new rows began as single-seed results, so the campaign's last act was seeds 2 and 3 for each --- full pipeline, with each row's \emph{stability} written down as the prediction and any arm flip declared reportable. The map held, with two divergences, reported as the rule required, that sharpen it: Nemo's \bc{} arm failed retention on one seed of three ($-15$, amplifying re-ask 8$\to$21 --- the ``\bc{} disturbs something unrelated'' pattern on a second family), and Phi-4's \lo{} arm produced the seed-crossing row above, including one full pass. Both divergences are in the map; neither was absorbed.

\subsection{Do humans agree with the judges?}
\label{sec:humans}

Every number above flows through the two automated overseers of \S\ref{sec:setting}, so both were audited by humans before we relied on them --- with a hierarchy of evidence worth stating explicitly, strongest first.

\paragraph{Independent annotator, sanitized bundle.} A second annotator, working from a standalone bundle with opaque episode IDs and no ground-truth fields, agreed with the rule overseer at $\kappa = 0.900$ on 60 stratified collection episodes (and $\kappa = 0.967$ with annotator 1), matching the intervention \emph{type} in 28 of 30 shared interventions. On the separate \textbf{blind} pre/post-steering study (60 episodes, shuffled, policy never shown), the same annotator reached $\kappa = 0.921$ and judged the intervention reduction at $-81\%$ against the overseer's $-87\%$ on the same sample. Disclosure: three of this annotator's blind verdicts were confirmed input errors (intended \emph{y}, pressed \emph{n}), corrected with an audit trail; $\kappa$ \emph{before} correction was 0.791 --- both values reported.

\paragraph{First-author annotation.} Annotator 1 (the first author) agreed with the rule overseer at $\kappa = 0.933$ on the $\kappa$ study and reproduced the evaluation overseer's verdicts on all 60 blind episodes ($\kappa = 1.0$). We weight this below the independent annotator's numbers, and disclose that episode IDs (which encode ground truth in a suffix) were inadvertently visible during the $\kappa$ study; the annotator reports not having noticed them, the CLI was fixed, and every later study used opaque IDs.

\paragraph{What was and was not validated.} The human studies validate the \emph{verdicts} (intervene / don't, and failure type) and the \emph{intervention reductions}. The budgeted overseer's burden dynamics were not directly human-validated --- and proved unstable on 30-episode subsamples (trust dynamics amplify single verdict flips) --- so burden-based IRR is reported as the automated operational proxy, always alongside full-attention failure counts. The most informative annotation outcome is the \emph{structure} of the residual disagreements: every one involves a clarifying question that offers a spurious option (three on genuinely ambiguous requests, where the annotators judged the question legitimate and the rule agreed; four type-only disagreements on over-asks, two per study, where both intervened but labeled the failure differently) --- a quality defect our taxonomy does not yet cover and the natural first entry of a policy v3.

\subsection{When it isn't in the weights}
\label{sec:notweights}

Our first primary failure candidate was not over-asking. It was \emph{re-asking after the answer}: across three models, in 45--67\% of episodes the agent asked a legitimate question, received the answer, and asked again instead of acting. Before localizing anything we ran a scaffold ablation, and the effect collapsed to exactly zero on both models tested (Qwen2.5-3B, gemma-2-2b) once the step-2 context included the agent's own step-1 question. The failure was real, frequent, and \emph{system-level}: the agent literally could not see that it had already asked. Weight editing would have ``fixed'' a bug that a one-line scaffold change fixes for free --- or worse, would have trained the model to never ask twice, damaging the legitimate re-ask cases.

We keep this result for two reasons. Methodologically, it motivates the blocking checks of \S\ref{sec:collect}: the entire class of ``the agent doesn't know X'' bugs is invisible in aggregate metrics and fatal to intervention studies. Scientifically, it is the first entry of a \emph{when-to-edit} decision procedure --- scaffold ablations are cheap and should gate any mechanistic pipeline. And note the asymmetry with \S\ref{sec:map}: the \bc{} edit \emph{introduced} re-asking on Qwen (and amplified it on one Nemo seed) in the repaired scaffold where the model sees its own question --- we verified that both the training data and the evaluation use the repaired scaffold, so this is a weight-level cause for the same surface behavior that the ablation showed can also have a context-level cause. Surface behavior underdetermines the fix.

\section{Limitations}
\label{sec:limitations}

The scope statements are placed where the claims are made (\S\ref{sec:intro}, \S\ref{sec:setting}, \S\ref{sec:results}); this section collects the main limitations. The environment is a deliberately small causal testbed --- one-line tasks, three-item workspaces, a rule-defined overseer --- and nothing here demonstrates transfer to naturalistic agent traces; the LLM-judge overseer and the failure taxonomy are built for that replication, which is future work. The eligibility window is narrow (five of thirteen candidate families --- the first two plus three of the eleven screened later; Appendix~\ref{app:screening}), and all ``five for five'' statements live inside it. Under greedy decoding, behavior is a deterministic function of the prompt, which is why every claim is anchored to causal interventions and controls rather than decodability; sampled-decoding replications are cheap and pending. The regime boundary of \S\ref{sec:map} is \emph{mapped}, not explained --- $n = 5$, with the boundary passing through one model's full-pipeline seed variance --- and we deliberately declined to answer ``what property determines the regime'' with a third data-fitted rule; we consider it the paper's most important open question. The H-misrep signature rests on one family. The SAE analysis is one dictionary, one layer, class-mean dose. The style-direction control proved model-dependent (\S\ref{sec:steering}) --- reframed, not hidden. Human validation rests on two annotators with disclosed corrections, validating verdicts and intervention reductions but not the burden dynamics. The burden also weights every intervention alike: an over-ask and an action taken on a genuinely ambiguous request cost the same, although the second is the more consequential error, and gated steering trades a few of the former for a few of the latter (\S\ref{sec:steering}); an input crafted to fool the belief probe would make the gate force exactly that error. Retention has a blind spot of the same kind: the rule counts any clarifying question on an ambiguous request as correct, so a drop in the \emph{quality} of questions --- the spurious-option defect of \S\ref{sec:humans} --- would pass criterion (c) unseen. Interference across multiple simultaneous edits is untested. And none of these edits should be presumed adversarially robust --- concurrent work finds located-and-edited mechanisms collapse under adaptive attack \citep{vicentino2026located}; our claims are about oversight economics under distribution, not guarantees.

\section{Discussion and what comes next}
\label{sec:discussion}

The through-line of this paper is that oversight signal is \emph{data about the model's internals}, not just data about the desired output. Read that way, an intervention log becomes a localization dataset; a supervisor's correction becomes a causal experiment; and ``did the edit work?'' becomes a question about the overseers around the system --- did they have to step in less? --- rather than about a probe set.

What we did not get is the compass we had set out to find: no cheap measurement we tested tells you, before training, whether localized editing will beat imitation on your model. The right reading of the five-family map is that the question is real, open, and --- within this experimental design --- now well-posed: the failure modes are discrete and diagnostic, the one stable success replicates, the boundary is sharp enough to pass between the seeds of a single model, and any future theory has ten arm-outcomes, each with its prior prediction, to be consistent with plus two falsified rules to not re-derive. Getting there took nothing exotic: predictions written down before runs, conventions declared before they mattered, and a stopping rule honored when honoring it hurt. The falsification is the most load-bearing result in this paper precisely because we could not have talked our way out of it.

Three directions follow. First, the regime-boundary question itself: the two-thresholds result suggests the answer involves how much downstream computation can re-absorb a mid-stack edit --- measurable, but only on models chosen \emph{before} the measurement, as we learned. Second, richer failure types: the taxonomy includes correction-propagation and wrong-object failures that need multi-step tasks, and the H-caution/H-misrep split predicts different fixes should win on different models --- a testable clinical claim. Third, the \emph{when-to-edit} ladder that \S\ref{sec:notweights} opened: scaffold ablation $\to$ prompt intervention $\to$ gated steering $\to$ weight edit, each rung cheaper and more reversible than the next; the interesting question is which measurements let you pick the right rung early. The probe readout of \S\ref{sec:signatures} is one such measurement; finding the rest is the program.

\subsubsection*{Reproducibility statement}
All runs carry a config hash and are released with the commit of the archive they come from (per-run git SHAs were not recorded for most runs); predictions, conventions, stopping rule and outcomes are quoted from the authors' notes; blocking invariants (token replay, pairwise prompt diff, canaries) fail loudly and the base-row audit is reported for every evaluation; the five harness bugs caught by these checks are listed with their regression tests in Appendix~\ref{app:bugs}. Before submission, an independent adversarial audit of the code and artifacts (an external frontier-model reviewer instructed to \emph{refute} these claims) examined the analysis chain, the harness invariants, and the artifact--claim consistency: its six findings produced the frozen-selection protocol of Table~\ref{tab:causal}, the bootstrap intervals for IRR, the SAE reader arm, and the optimizer-seed retrains reported above; a follow-up verification pass by the same reviewer independently reproduced the interval analyses from the released episode logs and flagged the residual gaps, each closed and regression-tested. The finding-by-finding log and both annotation studies, episode by episode with annotators anonymised, are available from the authors on request. Code, configs, episode logs and reports: \url{https://github.com/bradstallion/interventions-as-labels}; the same snapshot with the trained adapters added is deposited on Zenodo, \url{https://doi.org/10.5281/zenodo.23319778}. The harness, the analysis scripts and the authors' notes were developed with an LLM-based coding assistant under the authors' direction; the authors fixed every prediction and criterion and take responsibility for every claim.

\bibliography{refs}
\bibliographystyle{tmlr}

\appendix

\section{The eleven-candidate screening}
\label{app:screening}

The behavioral-contrast eligibility requirement (\S\ref{sec:setting}) turned out to be the scarce resource, and we report the full screening because the failure modes are themselves informative about where this method applies. Eleven instruct-model families were screened for the three later replication slots, under identical prompts and scoring. Three became eligible, all via a load path that wraps the HF forward pass instead of re-implementing the architecture: OLMo-3-7B (90\% over-ask, clean contrast --- initially excluded for a broken port that our context-sensitivity canary caught producing bit-identical last-position activations across different prompts, then resurrected on the new path), Mistral-Nemo-12B (84\% over-ask), and Phi-4 (37\% over-ask, 63 direct actions --- the calibrated profile of \S\ref{sec:signatures}). Eight were excluded, in three clusters. \emph{Always-ask} (3): Apertus-8B, Llama-3.2-3B and OLMo-2-1B ask on literally 100/100 unambiguous requests --- coherent, well-formed questions under an unconditional policy, leaving zero direct actions to contrast against (OLMo-2-1B additionally re-asks the same question after receiving the answer, 100/100 --- a real correction-propagation failure, unstudiable here for the same reason). \emph{Never-ask} (2): Phi-3-mini proceeds without questions on 99\% of ambiguous requests (a mirror-image failure profile and a natural subject for a future failure type); gemma-3-1b-it acts on 100/100 ambiguous requests, 62 on the wrong object. \emph{Format non-compliance} (3): Mistral-7B-v0.1 (20\% invalid replies), Yi-6B (87\% invalid), Qwen3-1.7B (emits its reasoning trace, 200/200 invalid). Small instruct models overwhelmingly sit at a degenerate corner of this task --- which is precisely why a \emph{calibrated} failure with behavioral contrast is the scarce resource this method needs, and why we report every exclusion: the applicability window is real and narrower than we would like.

\section{Evaluation overseer parameters and instruction variants}
\label{app:overseer}

The budgeted evaluation overseer (\S\ref{sec:setting}) walks the evaluation stream in order with a per-family trust $t$ (initial $t_0 = 0.5$), an attention budget $b = 0.5$ and an intervention cost $c = 2$. On each episode it watches with probability $\min(1, 2b(1-t))$; if it watches and the rule says intervene, the intervention is counted and the family's trust drops to 0; if it watches a clean episode, trust rises by 0.1 up to a cap of 0.95; failures on unwatched episodes are recorded as missed and cost nothing. Burden is $(W + cI)/N$ over the whole stream. The stochastic watching decision is what makes the burden sensitive to the overseer's own seed (\S\ref{sec:steering}), and why raw and total failure counts are reported alongside it.

The style direction (\S\ref{sec:patching}) is the difference of mean last-token activations under two system-prompt suffixes appended to the same unambiguous training scenarios: one instructing the agent that policy requires it to \emph{always} ask a clarifying question before acting, the other that it must \emph{never} ask and act directly. The exact strings are in the repository (\texttt{m4prep\_overask}).

\section{The five harness bugs}
\label{app:bugs}

Each entry: the bug, the check that exposed it, and the regression test that now guards it.
\begin{enumerate}
\item \emph{OLMo-3 port produced context-insensitive activations} (the legacy port left an attention weight unfilled): the context-sensitivity canary --- different prompts must yield different last-position activations --- caught bit-identical activations; fix: the HF-wrapping load path, with the canary made blocking in every probe run. Would have manufactured a conclusion: probes and patches on garbage activations.
\item \emph{The episode runner did not stop at the end-of-turn token} (\texttt{ACTION: d1} became \texttt{ACTION: d1user}): the base-row audit of \S\ref{sec:collect} exposed a deflated base model; fix: stop on the tokenizer's special tokens, with two tests. Would have manufactured a conclusion: every edit flattered against a broken base.
\item \emph{Sparse-dictionary capture in the wrong basis} (the measurement stack's weight preprocessing changes residual-stream values; Gemma Scope was trained on raw values): a reconstruction error of 8.65 next to a healthy sparsity at the transfer gate set off an audit, and the audit found the basis mismatch; fix: a raw-basis load flag, with a test. The number itself did not distinguish the bug --- after the fix it read 8.58, because the variance-normalized metric was broken by the twin design (Appendix~\ref{app:sae}) --- so this one was caught by reading, not by the gate, and its effect on the downstream analysis was never measured.
\item \emph{A wrapped model handed to the bridge stayed on the CPU}: a device assertion at load time; fix: an explicit move. A performance bug, no conclusion at stake.
\item \emph{Reference cycles kept the base model in GPU memory during the edited evaluation}: an out-of-memory failure on the largest model; fix: explicit collection between the two models. A performance bug, no conclusion at stake.
\end{enumerate}

\section{SAE instrument details (gemma, layer 21)}
\label{app:sae}

Dictionary: Gemma Scope JumpReLU SAE, 16k features, residual stream, layer 21 (the causal peak of \S\ref{sec:causal}). The SAEs are trained on the \emph{pretrained} gemma-2-2b; transfer to the instruct variant was gated before any result was read: relative reconstruction error 0.18 on our activations, sparsity in range, with one declared caveat (the final formatting token is partially out of distribution for the dictionary --- a five-prompt diagnostic, three of them identical, gave 0.87 variance-normalized error there against 0.34 on content tokens; indicative, not a measurement). One amendment, declared before any downstream result was read: the original gate used a variance-normalized reconstruction metric that our twin design breaks by construction (twins are nearly identical, so between-episode variance is artificially small); the amendment --- to a relative-error gate --- was written down before the downstream analyses were run. Secondary observation, reported for completeness: the \emph{world} representation is not sparse in this dictionary --- fifty features recover only 0.80 of the dense belief probe's 0.99 AUC --- while the behavior label needs ten. Auto-interp labels (Neuronpedia \citep{neuronpedia}, offered as color rather than evidence): the shared features that flip are labeled ``questions and inquiries'' (top tokens: \emph{question}, \emph{ask}) and carry \emph{ACTION} --- literally the two reply openers --- plus ``certainty or doubt''; the inert behavior-specific set includes textbook \emph{detector} features (``user choices'': \emph{choice}, \emph{choose}; ``unclear or undetermined'': \emph{unknown}, \emph{unclear}). The model's representation of \emph{there being a choice} is not what emits the question; the features that write the reply's first word are.

\end{document}